\documentclass{ws-book9x6}
\usepackage{ws-book-thm}   
\usepackage{ws-book-har}   
\usepackage[pdfpagelabels=false]{hyperref}  
\usepackage{here}

\title{Handwritten Historical Document Analysis, Recognition, and Retrieval -- State of the Art and Future Trends}      

\begin{document}


\begin{dedication}
{\large Lifelong learning for text retrieval and recognition in
historical handwritten document collections}\\
Lambert Schomaker\\
\medskip
To appear as chapter in book: \\
Handwritten Historical Document Analysis, Recognition, and Retrieval -- State of the Art and Future Trends\\
in the book series:\\
Series in Machine Perception and Artificial Intelligence\\
World Scientific\\
ISSN (print): 1793-0839
\end{dedication}







\setcounter{page}{1}
\setcounter{chapter}{9} 



\chapter[Lifelong learning for text retrieval and recognition in
historical handwritten document collections]{Lifelong learning for text
retrieval and recognition in historical handwritten document collections\\
{\small (Lambert Schomaker)}}\label{ch9}


\section{Introduction}\label{sec9.1}

Current developments in deep learning neural networks show remarkable
progress, also in document-analysis systems for historical manuscript
collections~\citep{Sudholt2018-ACF,Gurjar2018-LDR,Sudholt2016-PAD,Bluche2016e}.  
However, there are still many stumbling blocks and
recognition performances are still not at a level which matches the
expectations in user communities.  For scholars in the humanities, these
expectations are indeed high, as evidenced from the criticism that
large-scale digitization endeavors such as the Google Books project\footnote{\url{https://en.wikipedia.org/wiki/Google_Books}} 
drew~\citep{Chalmers2017}\footnote{Such opinions were raised quite vocally, e.g., 
during plenary discussions at
the Annual Seminar of the Consortium of European Research Libraries (CERL), Oslo, Norway,
October 28th, 2014}. This state of affairs is noteworthy, because optical character
recognition of machine-printed text - as opposed to handwritten manuscripts - 
yields performances that are already incredibly high from the point of view of
current handwriting-recognition research.  Practical tests reveal
character recognition rates from 68\% on early 20th century printed
newspapers to 99.8\% on modern material~\citep{Klijn2008}.  Lower
rates from 71\% to 98\% character recognition are mentioned for
newspapers from the period 1803-1954~\citep{Holley2009} who also reports that 
any performance below 90\% recognition accuracy would be considered 'poor'.  

If complaints on the accuracy of recognized text and its OCR-based
metadata are already so strong in the recognition of machine-printed
text, what can we expect from the user's reactions on current
handwriting recognition algorithms? It is clear that some reflection is
necessary in order to promote computer-based reading systems for opening
access to historical collections.  In this chapter a number of
considerations and experiences will be presented concerning the
development of the Monk~\cite{Zant2008,ISR2009,vanOosten2014,DesignConsSchomaker2015} 
e-Science service for historical documents at
the University of Groningen in the period 2008-2019.  This system aims
at supporting researchers in machine learning and scholars in the
humanities in doing research concerning the What, When \& Who questions:

\begin{itemize} 
\item 'What has been written?' (text recognition); 
\item 'When was it written?' (style-based dating of manuscripts); and 
\item 'Who wrote the document?' (writer identification). 
\end{itemize}

By adding labels at the page-description level, adding line transcriptions at the level of line-strip
images and adding zone labels for words and characters, the scholars create
a growing index to documents. At the same time, machine-learning researchers can
use the harvested $\{image,label\}$ tuples for training their methods.
Table~\ref{tab:keynumMonk} gives an overview of a number of relevant statistics.

\begin{table}[ht] \centering
\tbl{Key statistics representing the data present within the Monk system (July 2018).\hfill}
{\begin{tabular}{ll} \toprule
{\em Number of:} & (qty.) \\ \colrule
Institutions                \dotfill & 30 \\
Books, multi-page documents \dotfill & 567 \\
Page scans                  \dotfill & 152k \\
Line strips                 \dotfill & 273k \\
Zone (ROI) candidate images \dotfill & 700M \\
Lexical and shape classes   \dotfill & 147k \\
Disk storage                \dotfill & 120 TB \\
Files                       \dotfill & $1.3 \times 10^9$ \\
Human-labeled zones                  & ~ \\
~ words, characters, visual elements \dotfill & 900k \\ \botrule
\end{tabular}}
\label{tab:keynumMonk} 
\end{table}

\begin{figure}[h]
\centerline{\fbox{\includegraphics[width=10cm]{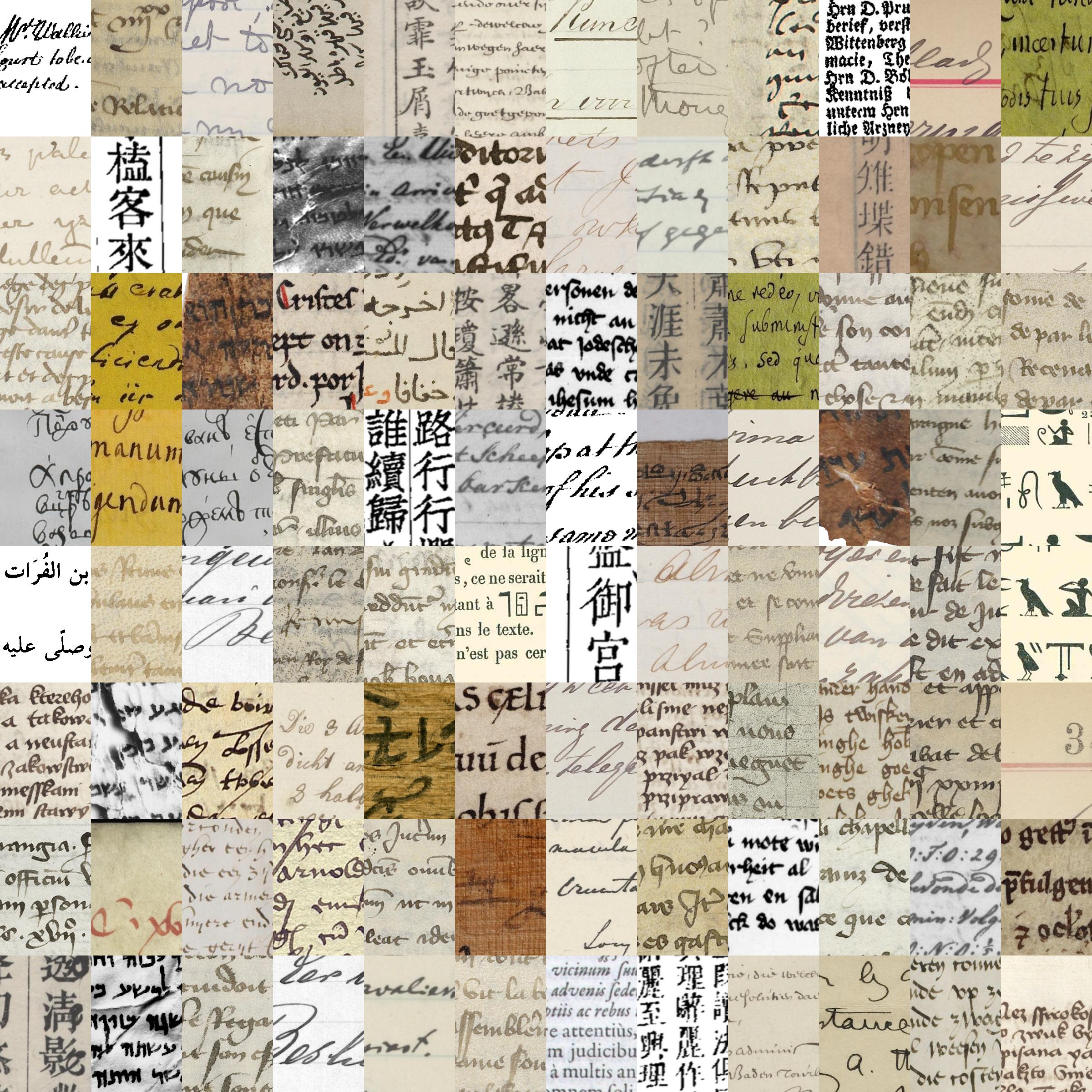}}}
\caption{Random samples from the collections currently on the Monk system. Apart from
handwritten manuscripts also some machine printed material is on the system. In the
philosophy of big-data methods, the classification algorithms should not be trained
or designed specifically for a particular style. While this is possible to a large extent, 
the wide range of image quality and layout particularities remains challenging.}
\label{fig:CollectionThumbs}
\end{figure}

In the context of the Monk system {\em Human labeled} means: 'an image zone, manually 
labeled as an original individual text item, possibly new to the system', or, alternatively, 
it may imply: 'recognizer-based labels that are confirmed by a human user on the basis 
of a ranked hit list provided by the system'. Figure~\ref{fig:CollectionThumbs}
gives an overview of the document styles, scripts, and image quality varieties
in the current collection on Monk.

%
%
\setlength{\fboxsep}{0pt}
\setlength{\fboxrule}{1pt}

During the development of this system, the following issues were encountered, concerning
the questions of the users and concerning the machine-learning approach to choose:

%
%
%

\begin{itemize}

\item Expectation management - How is a reading system positioned as a
technological tool in terms of the promised functionality and the
benefits to be expected for users, i.e., librarians, archivists and end
users?

\medskip

\item Usage scenarios - Does the institution basically desire 
an e-Book reader with a key-word search enhancement, is the goal 
to support users in creating edited digital books from handwritten
manuscripts or is the goal to just convert images of paper documents 
to encoded text for further processing?

\medskip

\item Technical realization - Is the reading system offered as a
stand-alone computer tool, running on the the PC of the end user, or
running on the infrastructure of a library
institution under local supervision, or is it constructed as a
'cloud'-based service, somewhere on Internet, with support from remote,
possibly anonymous but expert-level staff?

\medskip

\item Work-flow concept - Does the archive intend to perform a
single-shot conversion of a given collection (from pixels to encoded
text) or do the users realize that continuous efforts in quality control will
be required, including labeling by experts or volunteers over
a prolonged period of time?

\medskip

\item Quality and quantity of the material - Will there be a preselection
of materials and what are the criteria? What types of material are
present: books, diaries/journals, shoe boxes with letters in their
original envelope, written by diverse writers?

\medskip

\item Industrialization and scalability - Who is responsible for image
preprocessing and layout analysis? A folder with 2000 scanned raw
unlabeled images of papyrus texts is not well comparable to an academic
benchmark test for machine learning that is preprocessed, packaged, labeled and
prepared for k-fold evaluation experiments~\citep{MNIST1998,IAM2002,GW2012}. 
In real-world applications, collections have a wide
variety of document and layout formats. How to handle the consequences
of success: "If it works, can we process several hundreds of such
collections within a year?"

\medskip

\item Human effort - Who will be responsible for quality control of the meta-data input,
and the linguistic quality of the labeling process? If users perform an inconsistent
labeling, they may blame the system if recognition performances are lower than expected.

\medskip

\item Algorithms - Finally, after having listened to signals from user communities,
there is the last consideration, which mostly is the starting point for many in our research
domain: The selection of the machine-learning
methodology used. How to choose between word-spotting~\citep{Rath2007}, word-based
recognition~\citep{Zant2008} and character-stream based handwritten-text recognition (HTR)~\citep{Sanchez2013}?
Even here, realistic and pragmatic considerations need to be taken into account, that are
insufficiently addressed by designers of machine-learning methods.
Will it be possible to enjoy improvements in algorithms, over time? 
How to select such methods? Are current methods in
deep learning usable in practical application settings?

\end{itemize}

\medskip

The answer to such questions is not easy. When maintaining a large-scale
e-Science repository of handwritten document collections, it 
quickly becomes apparent that at least 'four Vs of Big Data' play a role
here: {\bf Volume}, {\bf Velocity}, {\bf Variety} and {\bf Veracity}. 
The {\bf volume} is indeed large and may be in the
hundreds of thousands of page scans. This has consequences for storage
but also for computation: (re)training efforts will pose a significant
load on the infrastructure. Data sets can be much too large to fit in 
memory, especially considering the fact that training processes for
multiple collections will run in parallel. {\bf Velocity} pertains to
the rate at which new scanned pages are entering the system as well as
the rate at which labels are being produced by the users. New labeling
may refute existing labeling. Since neural networks cannot unlearn 
individual input/output pairs, new insights or corrections by users 
usually necessitates a new training from start. {\bf Variety} plays a
role in image-quality preprocessing. Although an 'end-to-end' training
is advocated these days, the generalization from, e.g., a parchment
training set to a papyrus-based test set will be highly limited.
A proper preprocessing increases the reusability of samples that
are homogenized in their visual appearance over multiple diverse training 
contexts. Variety also plays a dominant role at a level where current
deep learning does not provide solutions as yet: Layout structure diversity.
Curvilinear line shapes, tabular text objects and marginal notes often 
require collection-specific layout analysis. Finally, writing style variety
and linguistic variations need to be handled. As regards {\bf Veracity}, it
should be noted that the machine-learning assumption that 'ground
truth is a rock solid factum' cannot be held in a live system. In many cases,
a coherent labeling systematics needs to be invented by humanities
researchers, on the spot: This goal may have been the whole purpose of the 
digitization effort for the manuscript, in the first place.

\medskip

In this chapter, the evolution of the trainable search engine for
handwritten historical collection, Monk will be used to illustrate the
manner in which these considerations were addressed.  In this process, a
number of phenomena in machine learning were encountered that are
hitherto not handled in great detail in the common benchmark-oriented
literature but which will become apparent under large-scale, open domain
conditions.

\section{Expectation management} 

Both in the Google Books~\footnote{\url{https://en.wikipedia.org/wiki/Google_Books}} 
project and in projects aimed at digitizing
vast amounts of handwritten material from a wide range of international
scripts and historical periods, it is essential to create realistic
expectations in end users.  The end users can be archivists and
librarians, for whom correctness of metadata and core text data is very
important.  End users have their own expectations, which are
concentrated in the areas of legibility and the reading experience, i.e.,
usability aspects, as well as accuracy of text recognition~\citep{ISR2009}.
Increasingly, there are expectations as regards multimodality, the 
recognition of the pictorial information next to the handwriting, and 
the possibility of (semi-automatic) creation of hyperlinks to semantic 
databases~\citep{WeberEtAl2018}.

%

\begin{figure}[h]
\centerline{\includegraphics[width=10cm]{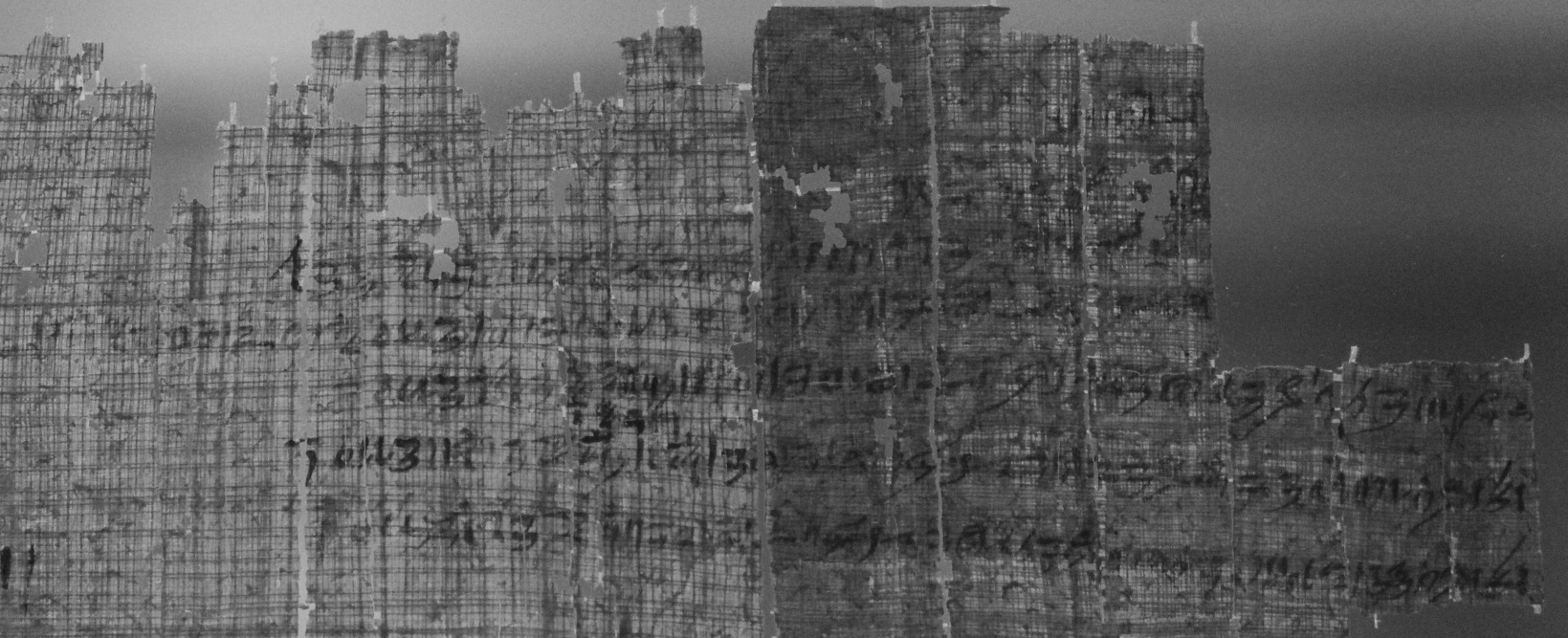}}
\caption{An example of hieratic script with papyrus in two types of aging within the
same text column and weak contrast differences between ink trace and papyrus fibers.}
\label{fig:Hieratic}
\end{figure}

\begin{figure}[h]
\centerline{\includegraphics[width=10cm]{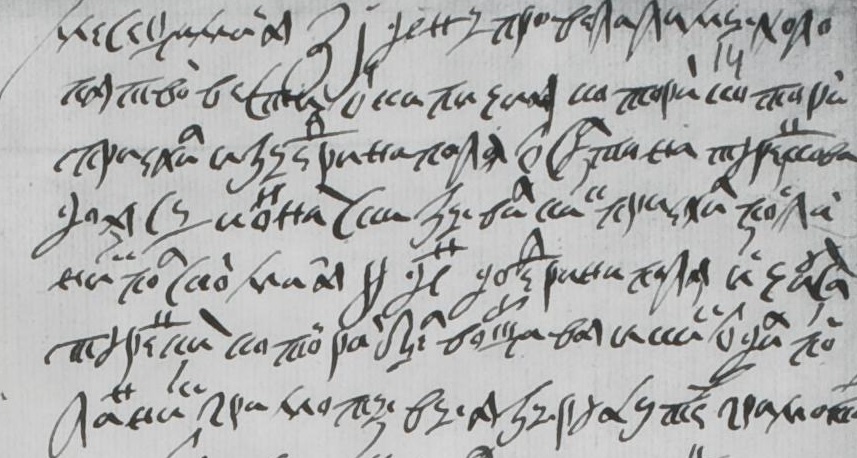}}
\caption{A 17th century Russian example of an exuberant style with isolated and connected-cursive 
elements as well as overlap of ascenders and descenders between successive lines.}
\label{fig:Russian}
\end{figure}

Inspection of the handwriting-recognition literature will lead to
unrealistic expectations. From a fundamental-science perspective,
standard benchmark tests are clearly necessary. However, they also
introduce a bias as well as a reduction in the diversity of data that
are handled in literature, in comparison to real-world applications. 
The standard data sets, especially MNIST~\citep{MNIST1998},
but also the George Washington (GW) data set~\citep{GW2012} have been introduced quite
a number of years ago, with generations of PhD students trying to
improve the recognition rate of the previous cohort, which is possible
due to the vast accumulation of knowledge and skills concerning
precisely these data sets (but not other unseen data sets).  Even the
IAM data set~\citep{IAM2002}, which is very realistic and challenging due to the presence of
multiple writers, is of limited use as a predictor for the performance
of an algorithm on historical material. The handwriting style
and image quality concerns contemporary mixed-cursive and cursive handwriting. 
Recognition results obtained for such data sets are meaningless for an
archivist with a medieval collection, a collection hieratic script on
papyrus (Figure~\ref{fig:Hieratic}), or a collection of Russian 17th century pamphlets in an
exuberant cursive style with large ascenders and descenders (Figure~\ref{fig:Russian}).  


\begin{figure}[h]
\centerline{\includegraphics[width=10cm]{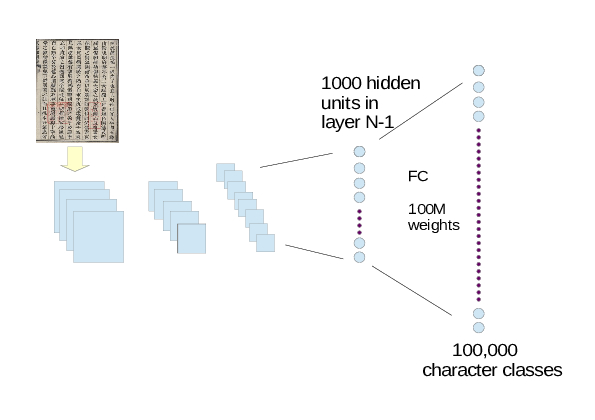}}
\caption{In case of a very large number of output classes, the notion of
a one-hot target output vector representing the classes becomes untenable.
Handcrafted attribute representations need to be designed for the final layer,
reducing its dimensionality~\citep{ShengOpenSet2018}.} 
\label{fig:ChineseCNN}
\end{figure}

The goal
of performing tests under acceptable theoretic conditions of
independently sampled and identically distributed statistical properties
is laudable and necessary, but it is not enough.  Real collections
violate both the conditions of stationarity and ergodicity which are
needed for successful machine learning.  A writer of a journal (diary) will
usually start out with high ambitions and a clean style, in eloquent full sentences.
However, as time progresses, this may evolve into sloppily produced entries containing 
more and more abbreviations, marginal notes in all orientations, and other
idiosyncrasies (doodles) that were not present in the first pages.  Both
at the level of shape and at the level of language, the source processes
are not stationary.  Ergodicity, the requirement for a stochastic
process that its statistical distribution can be estimated (read:
because there is a single, underlying signal source) is also often
questionable.  A single document may be produced by multiple scribes
(read: different writing-style signal sources) and the linguistic
content may vary from chapter to chapter to such a degree that one
cannot consider the overall process ergodic: it is based on a diverse
set of stochastic language generators.  Such considerations seem to
point to the use of machine-learning models that are trained and
applied, {\em conditional} to a known aspect of the signal source, e.g.,
"we are in a Chinese section of the series of page scans", which
indicates that a deep-learning model trained on Chinese characters may
be considered as more appropriate than other style models.  More
formally, if $p(x|M1,S1) > p(x|M2,S2)$, then $p(C|x,M1) > p(C|x,M2)$,
for samples $x$, a sample class $C$, two styles $S_{i}$ and
corresponding trained models $M_{i},i=1,2$. In words, if the data
are likely to be generated by model M1 if the source is in style S1, the 
probability is higher that a correct class label C is 
computed by model M1 rather than by model M2. An alternative approach
would, e.g., require the assumption that there is enough training data for all
possible scripts and classes such that a true 'omniscript' neural network
can be trained.  In this second, ambitious approach it is assumed,
unrealistically, that a) enough data are present to bypass the problems
of non-stationarity and non-ergodicity and that b) the resulting model
would not suffer from the competition between all the classes, i.e., the
complete set of international scripts.  Please note that the Chinese
language alone has a dictionary set (Ytizi Zydian) of 106.203 character shapes.  
Although
other languages have smaller character sets, the mutual competition is
bound to create risks.  If whole-word classification is performed, the
number of classes will be equal to the sum of the word lexicon sizes of
all involved styles.  For the case of Chinese alone, a convolutional
neural network with a 'normal' dimensionality of a thousand units in
pre-final layer N-1 and 100k output units in 'one-hot' configuration 
would require a fully connected layer with more than 100
million weights~(Figure~\ref{fig:ChineseCNN}).  At this number of coefficients it is unlikely that
the 'dropout trick'~\citep{Dropout2012} will alleviate this situation.  


\begin{figure}[h]
\centerline{\includegraphics[width=10cm]{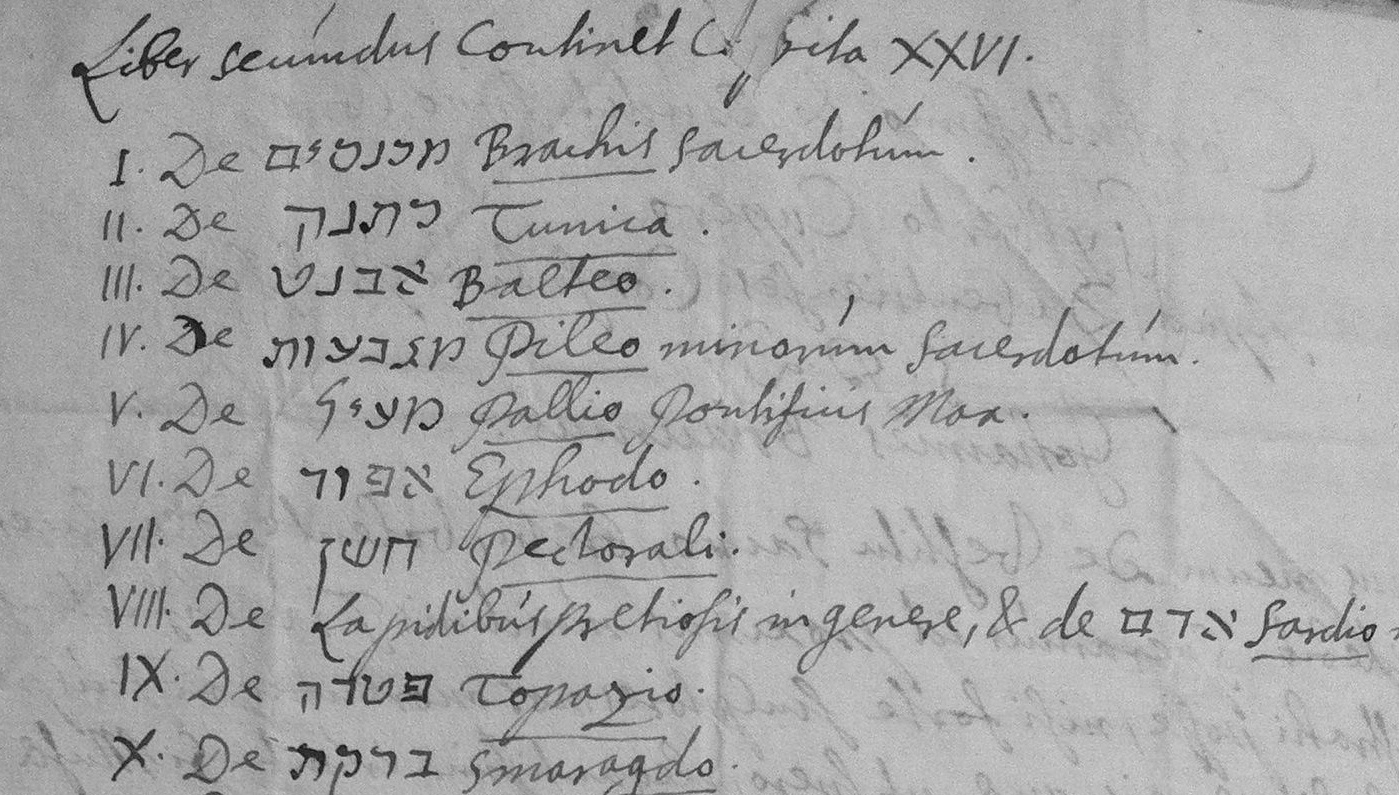}}
\centerline{\includegraphics[width=10cm]{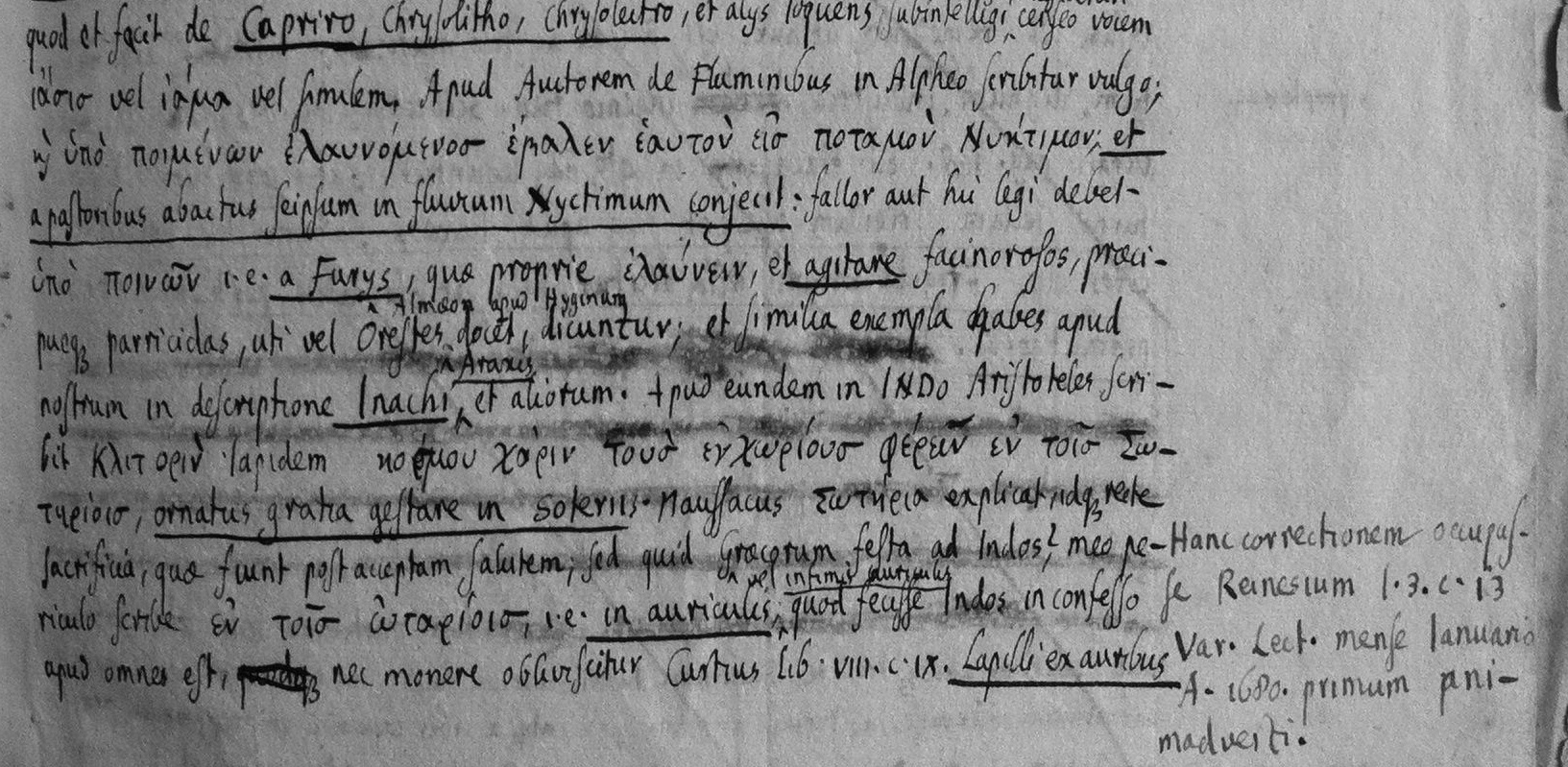}}
\caption{Examples of mixed Latin, Greek and Hebrew text within 17th century European scholarly letters.
In the same collection, also correspondence letters containing French and other languages can be found next
to fragments of classical language and script.}
\label{fig:ScholarlyLanguageDiversity}
\end{figure}

With such a large number of incoming weights to a neuron, the collinearity problem
entails that there are too many equipotential solutions for realizing
a target (output) activation value, turning training into an ill-posed problem,
i.e., a problem for which not one or, alternatively many solutions exist.
It this example, the estimation problem is so massive that it can be argued 
that even in the absence of a collinearity problem, the underlying 
densities (the shapes of the underlying manifolds) need
to be covered in sufficient detail by giving the neural network a sufficient number
of training examples. Considering a 
lenient rule-of-thumb to have 5 samples (I/O pairs) per coefficient 
this would require half a billion training samples, disregarding 
the number of weights in the earlier layers in that network. If one
would argue, that with a dropout probability of, say 0.8, the net number of 
coefficients is 20 million weights, this still implies that 100 million
labeled samples would be needed according to the rule-of-thumb, 
a quantity that few research groups have 
at their disposal. With some additional (human) thinking, other representations
than 100k-dimensional one-hot vectors can be used to solve the large
set classification problem.
We have proposed a solution based on
attribute learning for the case of very large character sets~\citep{ShengOpenSet2018}.
Within the Chinese character set, convenient 
attribute representations concerning the presence of radicals but
also the order of strokes (e.g., the Wubi Xing method~\footnote{\url{https://en.wikipedia.org/wiki/Wubi_method}}).
Similarly, for Western texts, an attribute method was introduced, PHOCNet~\citep{Sudholt2016-PAD}.
While very useful, such useful methods are highly script and/or language specific, thereby
blocking the ambition of an 'omniscript' approach. If it is necessary
to improve the recognition rates by using linguistic statistics, this 
introduces yet another condition that needs to be determined for processing
a given image region. An e-Science service for handwriting recognition will 
need to be able to handle a wide variety of scripts and languages. Sometimes,
it will not even be possible to flag individual pages with a code for script
style and language condition. The 17th century Cuper-Braun~\footnote{The Cuper-Braun collection was kindly provided
by dr. J. Touber} collection in Monk,
concerns a series of European scholarly letters by different writers. 
They write in a multitude of languages, switching from Latin to French, 
interjecting the text with phrases in Greek and Hebrew (Figure~\ref{fig:ScholarlyLanguageDiversity}).


Therefore, in any case, some form of modular classification approach appears
to be more practical when it comes to historical manuscripts.  As a
consequence, dedicated training will be necessary on most of the
ingested digital documents.  In some cases, it will even mean that
separate models are necessary for papyrus-based and for parchment-based
documents of a particular writing style, due to the clear differences in
the ink deposition and absorption process for the two materials. See Figure~\ref{fig:Hieratic} for 
an example of problematic pre-processing requirements for hieratic script on papyrus. 

In light of the problems lined out thus far, it may be clear that the
concept of an {\em AI black box} that will produce a perfect Microsoft Word text file
when given a scanned image of historical text is utterly unrealistic. Consequently,
the conclusion was to inform potential Monk system users according to the following
notes:

\begin{itemize}
\item Don't promise perfection
\item Don't promise exhaustive coverage
\item Don't promise that it is a single conversion step
\item Make clear that labeling will be necessary along the digital life cycle of the collection
\end{itemize}

\section{Deep learning}

The successes of deep learning are based on a number of favorable
conditions that are lacking in the context of historical document
analysis.  The problem of the archivist consists of a pile of scanned
images, with an image quality that is not guaranteed to be amenable to
optical character recognition or handwritten text recognition
beforehand.  The language may be known, but that does not mean that an
appropriate lexicon exists.  The same holds for other linguistic
resources.  The character shapes,allographs are usually unknown, or a
variant of a known style and no labeled data for machine learning is
present.  In fact, the whole purpose of the digitization is to get at
encoded text, given the pixels in the image.  What can be done to enjoy
the benefits of current machine-learning technology? One possible
approach is proposed with the Transkribus~\citep{Transkribus2017} platform 
within the READ project. 
Scholars are required to label a sufficient amount of pages, by
transcribing them at the line level.  Typically 70 to 100 transcribed
pages are considered to be necessary.  This approach has limitations. 
For instance, what is the optimal set of pages to be used as the training set for a
larger collection? Will it concern a single run of human transcription activity?
As discussed above, starting with the first pages of
a handwritten document is risky due to the intrinsic non stationarity. 
Secondly, the benefits of the labeling will only become apparent after
the investment of human labeling has been done.  In the design of the
Monk system, an alternative approach is proposed, where pattern
recognition starts upon the input of the first word label.  From an
over-segmented data set, hit lists are generated of 'mined' word zones
and presented to the user, for confirmation.  The user may choose to
label words on the basis of a given line of text, thereby enlarging the
covered lexicon. This type of process was dubbed {\em 'widening'}.  
Alternatively, the user may choose to
confirm labels in the hit list for a pattern, thereby improving the training set 
for that particular class: {\em 'deepening'}.  In labeling the hit lists, also
additional catches of well-segmented words can be labeled, adding to the
lexicon of known patterns.  In this framework, useful computations are
applied at the earliest moment in time.  However, not all
machine-learning methods are suitable in such an approach, as will be detailed
in the next section.

\section{The ball-park principle} 

In fact, while a label-agglomeration process is evolving over time, the
system passes different ball parks, each with its own most suitable
machine-learning approaches. 

\medskip
\noindent
{\bf No labels (zero labels)}.  - In case no labels are available, it
may be possible to rely on a pre-trained recognizer from the nearest
handwriting style and a comparable lexicon.  Additional training can
then be applied for transfer learning.  A particularly interesting
solution is provided by machine-learning approaches based on attributes. 
Attributes are symbolic tokens that can be used to provide a text
hypothesis in what basically constitutes are rule-based computation~\citep{Sudholt2016-PAD,ShengOpenSet2018}
known as zero-shot learning~\citep{ZeroShot2008}.  Please note that in a
system such as Monk, which immediately exploits the arrival of new
labels, the underlying class of patterns can directly be trained.  In
this manner, a secondary classification method, B, is kickstarted by the
earlier applied primary method, A, which may be not optimal but still 
able to handle the zero-labels condition.

\medskip
\noindent
{\bf One label}.  - In case a single label is present, we enter the ball
park of nearest-neighbor (1NN) classification.  The labeled instance can
be used for a data mining process, as in word spotting.  Some
appropriate distance measure is needed for comparing feature vectors. 
Such feature vectors can be dedicated (designed) or be trained, as is
the case for embedded features tapped from some layer in an existing
pre-trained neural network~\citep{TappedFeatures2018}.  The resulting hit list can be
presented to human users for confirmation.  Upon entering a few labels,
the system enters the next ball park. 

\medskip
\noindent
{\bf One to five labels}.  With a few labels, some methods are able to
produce rudimentary models.  It becomes possible to compute standard
deviations of features, enabling the use of Bayesian modeling.  Instead
of 1NN, kNN can be applied but it is much more attractive to use
nearest-mean (nearest centroid) classification, because this reduces the
computation load.  After all, in a data mining context, if the reference
set is large and the pool of patterns to be mined is huge, computational
costs are too high.  Taking, again, the extremal example of the Chinese
dictionary set of 106k characters, the presence of 5 examples per class
would require 530k times N comparisons, N being the huge number of
candidate patterns in the data-mining pool.

\medskip
\noindent
{\bf Twenty to hundred labels}.  - As the number of labeled instances
increases, more powerful classification methods can be used.  For
instance, with 50 positive examples of a class and an appropriate number
of counter examples, support-vector machines (SVMs) can be trained for a
more advanced form of targeted (specialized) pattern mining in the pool
of unknown objects. 

\medskip
\noindent
{\bf More than one hundred labels}.  - Here, finally, a level of
coverage is obtained which allows training of contemporary deep-learning
methods.  However, this is only sensible if there are enough classes,
such that the total number of patterns is in balance with the number of
weights (model coefficients) in the classifier.  This amount of labeling
is just a starting point.  As is long known in professional OCR
classifiers for machine print, each class requires thousands of
examples. Similarly, for handwritten digits large data sets are 
currently being collected with 830k samples~\citep{Uchida2016}. 


\begin{figure}[h]
\centerline{\includegraphics[width=12cm]{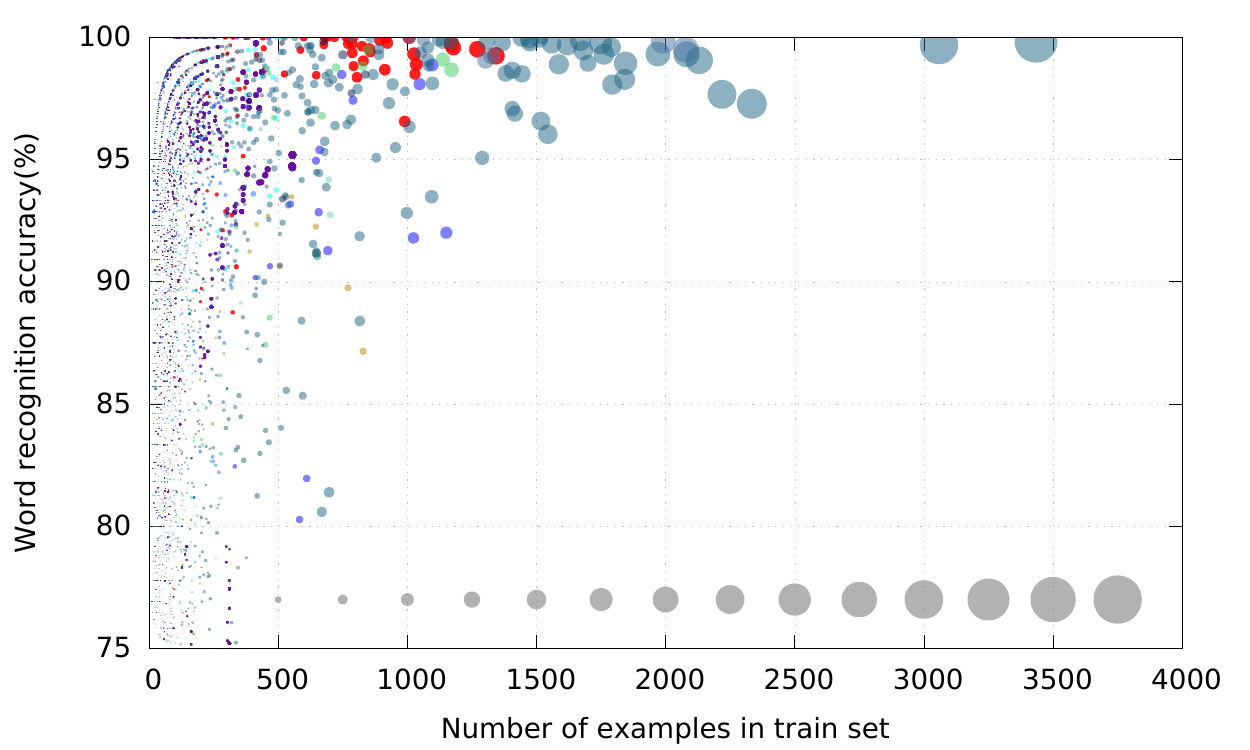}}
\caption{Distribution of recognition accuracy as a function of labels, 
for a wide range of documents, script types and languages. Word classes
will be moving towards the right and up over time, until a performance
asymptote is reached. Some classes already have a decent performance
with less than one hundred labels. On the other hand, some difficult
shape classes remain just below 100\% accuracy. The color of a circle 
denotes the manuscript group. The radius of the circle represents the
size of the test set and is an indicator of the harvest.}
\label{fig:PerformanceDensity}
\end{figure}

\begin{figure}[h]
\centerline{\includegraphics[width=10cm]{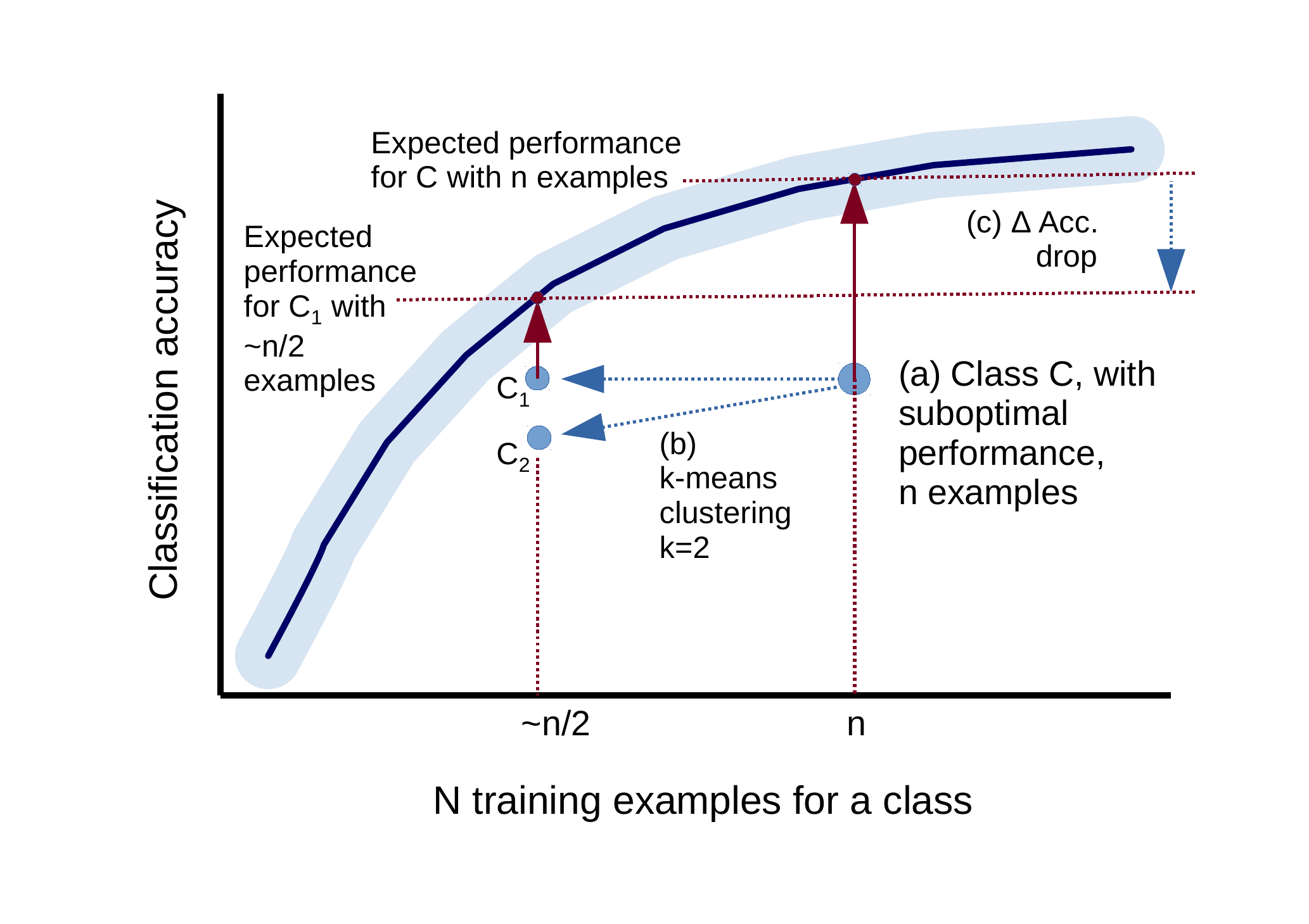}}
\caption{Schematic representation of the effects of 'precise' modeling
by splitting the training set for a class in style-based subclasses. (a)
a class $C$ shows suboptimal performance. (b) The idea is to use, e.g., k-means
clustering with k=2 to zoom in on the shape differences and have two specialized models $C_{1}$ and $C_{2}$.
(c) As a result of the smaller training sets, there will be a considerable drop
in expected performance for the specialized models. On the basis of the large set
of reference experiments, it is statistically questionable whether the supposed
exactness of the models can counteract the large drop in accuracy due 
to the loss of examples.}
\label{fig:Misguided-precision}
\end{figure}

As Figure~\ref{fig:PerformanceDensity} shows (upper left), 
the word (or pattern) accuracy performance of single classes may approach
100\% recognition with thousands of examples, but not in all classes.  
Such scatter plots provide a deeper insight in where the friction towards
a higher performance resides. Consider, for instance, the case
were a class is performing suboptimally. Following 'old-school'
precision thinking, the tendency would be to use, e.g., k-means
clustering to separate style variants in the training set for
that class, thereby zooming in to a more precise modeling of
allographs and increasing the number of shape classes. However, 
as Figure~\ref{fig:PerformanceDensity} shows,
this would entail an approximate halving of the number of examples
from $n$ to $\approx n/2$
corresponding to a large jump to the left on the horizontal axis,
towards a point where the estimated expected accuracy (vertical axis)
will be much lower. Indeed, we found that the performance will usually
decrease after such an 'increased model-precision operation'. It is much 
easier to increase the performance by adding labels than to improve a given
classifier on the basis of a reduced set of examples.
There is no data like more data! Apparently, if the recognizer method is powerful
enough, it is better to have mixed densities for style variants
in one classification model, than have, e.g., two specialized models 
where each is trained on half the size of the original set of examples (Figure~\ref{fig:Misguided-precision}).
We will only know whether the specalisation in modeling was worthwhile 
after our lifelong-learning engine has been able to harvest the original large
number of labeled samples for each of the subclasses $C_{1}$ and $C_{2}$ that was 
present in the lumped set $C$ at the onset, i.e., $n$ examples.

In any case, by the time the final ball park of abundant labeling has been reached, 
a solid ground is present for constructing an index of a document.  If linguistic models
are present, an attempt can be made towards full-page transcription.
Infrequent shapes and classes still pose a risk at this stage.  It may
be desirable to replace uncertain recognition results with an ellipsis
{...} instead of presenting uncertain text hypotheses. Presenting
post-processed results introduce a new type of problem, especially if a
recognized text is fluently legible, but unfortunately not what has been
written.  The user interface needs to allow the users to compare
recognized results and handwriting one on one.  If this particular risk
is not communicated to users, their disappointement with the system if
they detect such output may be larger than necessary.

\section{Technical realization}

Is the reading system offered as a stand-alone computer tool, running on
the infrastructure of a library institution under local supervision, or
is it constructed as a 'cloud'-based service, somewhere on Internet,
with support from remote, possibly anonymous but expert-level staff?

\subsection{Work flow} 

Does the archive intend to perform a single-shot conversion of a given
collection (from pixels to encoded text) or do they realize that
continuous efforts in quality control will be required?

\subsection{Quality and quantity of material}

Will the be a preselection of materials and what are the criteria? What
types of material are present: books, diaries/journals, shoe boxes with
letters in their original envelope, written by diverse writers?

\subsection{Industrialization and scalability}

Who is responsible for image preprocessing and layout analysis? A folder
with 2300 scanned raw unlabeled images of papyrus texts is not well
comparable to an academic benchmark test for machine learning that is
preprocessed, packaged and prepared for k-fold evaluation experiments. 
How to handle the bad luck of success: "It works, now we need to process
several hundreds of such collections within a year."

\subsection{Human effort} 

Who will be responsible for meta-data input, labeling, linguistic
quality evaluation?

%
%



\subsection{Algorithms}

As the last consideration, the machine-learning methodology used: Will
it be possible to enjoy improvements in algorithms, over time? How to
select such method? Are current methods in deep learning usable in
practical application settings?

\subsection{Object of recognition: Whole-word approaches}

For an e-Science service in historical document analysis, it is
important to strive for generic solutions wherever possible, given the
large scale and huge diversity.  Traditional speech and handwriting
recognition usually rely strongly on linguistic resources.  These are
usually not present for unique documents that just enter the digital
stage of their life cycle.  The assumptions for 'optical character
recognition' as 'classification of sequences of character classes' are
not valid for a wide range of documents.  At the same time it was clear
in the period 2000-2008 that general image classification was making
large leaps forward~\citep{Rath2007}.  In the document-analysis community,
word-spotting techniques were being proposed.  We found that
biologically inspired neural networks provided very high
word-classification rates on connected-cursive script~\citep{Zant2008}. 
As a consequence, a word-based classification approach was opted for in
Monk. 

\subsection{Processing pipeline}

\medskip
\noindent
{\bf Ingest}.  - The processing pipeline starts with the {\em ingest}
stage of a collection.  It is verified that the users will have a
long-term interest in the document, there is a responsible local
'editor' who instructs a group of volunteering labelers.  The material
needs to be sufficiently homogeneous, sufficiently large: much more than
a hundred page scans, and of a manageable image quality.  A choice is
made for the basic text object on the scans that will be the target of
recognition.  This often requires a split in separate recto/verso images
and a subsequent layout analysis to identify the major text columns in
the document.  Over time, a growing library of software scripts is
collected such that new collections require just a variation on a known
preprocessing scheme.  This stage usually requires some additional
programming using image processing tools.  For instance, for a
multi-spectral collection, a flattened image version that is optimal for
ink/paper separation should be produced.  The initial data transfer is
realized using ftp,transfer web sites or hard disks sent via surface
mail.  We have experimented with allowing an upload of individual
unorganized raw scans but this does not bring much to either party. 
Such isolated small projects do not benefit from the large scale of
data-science approaches and the isolated material gives little chances
to act as a multiplier for the solid coverage of a hitherto unseen
script style. 

\medskip
\noindent
{\bf Line segmentation}.  - Using horizontal ink-density estimates which
are low-pass filtered, an automatic segmentation into lines is
attempted.  This is first done on a random subset of the scans, in order
to find a proper parameter setting.  Perfection is difficult to realize
with a considerable portion of submitted collections.  For difficult
cases, curvilinear segmentation methods such as seam carving are
applied~\citep{SurintaLineSegAstar2017,ChandaLineSegmentation2018}. 
Current approaches in machine learning favor end-to-end classification
that starts with the original color image.  However, the curvilinearly
cropped text line requires an artificial background.  Replacement by
pure white pixels is not desirable, because it will introduce sharp
non-text edges along the curvilinear cut.  In
\citet{ChandaLineSegmentation2018} an attempt is realized to construct
an artificial background with the texture and color properties of the
original image, replacing undesirable ascenders and descenders of the
surrounding text lines.  Even such an advanced method will not solve all
possible problems.  Collections with marginal notes and post-hoc
corrections will be intrinsically unsolvable without some form of human
intervention such as providing manual segmentations for deviant objects
in the image.  If the goal is a mere indexing (as opposed to full
transcription) the attribution of a word to a particular line is less
important.  In many applications it will already be a great benefit if a
target key word can be found and marked on the page, disregarding its
membership to a line. 

\medskip
\noindent
{\bf Word-zone candidates by over segmentation}.  - Using vertical
ink-density estimates, candidate word zones are segmented which are of
variable overlapping size.  Widely spaced connected-cursive text is
ideal: The number of word zones will be limited and wide white spaces
prevent the occurrence of multi-word image crops.  At the other end of
the spectrum would be a faded typewritten text where even a single
letter may consist of multiple connected components.  Also here, some
experimentation with a random subset of page columns is required.  The
segmentation is facultative.  If at a later stage other word-zones are
added by another word-segmentation algorithm, the resulting candidates
are just added to the pool of word-zone candidates. 

\medskip
\noindent
{\bf Word recognition}.  - At this stage, the document will enter the
autonomous continuous learning cycle.  Word classification tools will be
applied to it, generating word-hypothesis lists for each word-zone
candidate.  The text hypotheses are added to a large raw {\rm Index}. 
In the training of the word classifiers, data augmentation is performed
using random elastic morphing~\cite{BulacuAugmentation2009}, which 
is necessary for all classes that have less than about 20 examples.

\medskip
\noindent
{\bf Word ranking: Hit list generation}.  As described
elsewhere~\citep{vanOosten2014} we found that using the likelihood
estimates generated by recognition engines, for instance the maximum a
posterior likelihood $p(C|x)$ does not by necessity yield believable,
intuitive hit lists.  The task of separating an instance class from
competing candidate classes, i.e., recognition, is another function than
ranking, where one wants instances to appear sorted at the top of a hit
list, with a strong similarity to a canonical model, i.e., optimizing
the reverse probability, $p(x|C)$.  Therefore, a method is used that is
optimal for ranking, yielding hit lists with understandable positive
results and even understandable errors.

\medskip
\noindent
{\bf Presentation of recognition results to the user: per line, or per
hit list for lexical words}.  At this stage, the user will be able to
confirm text hypotheses or enter new word labels.  This is detected by
the event handler of the system, which will add a recomputation request
for the class models at hand.  This in turn will lead to re-recognition
batch jobs, which generates new, improved text hypotheses, usually
within 24 hours.  At this point, the system will thus return to the
Word-recognition stage, closing an iterative loop we have dubbed the
'Fahrkunst' principle.  Closing of the loop to realize an autonomously
learning system was realized in the summer of 2009.  Controlling the
feedback loop concerns the manual adjustment of computation policy
parameters: Which books to focus the computation on, with which amount
of delay time per cycle.  A distinction is made between cold and hot
books in the collection.  Hot books are those where the user community
has a great interest and displays a fruitful labeling harvest. 

\medskip
\noindent
{\bf Construction of alphabetic word lists and provisional
transcriptions}.  Finally, in this on-going process, there is a periodic
selection of confirmed text hypotheses and manual transcriptions that is
presented on internet as static files.  A mechanism for downloading
indices is also present, using download licenses with a limited time
validity for end users.  Together with end users, a 'work list' is
defined for this type of objects.  This part of the system is becoming
more and more important and a large diversity of export formats is
expected to be relevant at this level.  For handwriting, the Alto
standard is not optimal, but the Page XML format by the Prima group is
an example of a usable output format for a range of further applications
beyond this stage. 

\section{Performance}

Although performance evaluation for recognition are straightforward, the
same cannot be said of retrieval engines.  The usual performance
evaluation in machine learning assumes a train set, validation set and
test set, all with labeled raw-data instances.  Commonly, k-fold
evaluation is performed to obtain reliable performance indicators in
terms of accuracy, precision and recall.  For the labeled portion of the
data in a large document analysis system, a similar periodic evaluation
can be applied to select optimal classifiers per script style, and so
forth.  One might think that the recall performance is a good predictor
for the actual recall capability of a classifier in open-ended data
mining.  However, this is not straightforward while being in the middle
of a continuous learning process.  We do not know how many instances of
a particular class are out there, in the mining pool? If an asymptote is
reached in the accuracy for a particular class, does that mean that the
pattern is 'mined out', i.e., that no more instances of that class exist
in the pool, or, alternatively, is it the case that there is a
particular problem at the level of shape representation and
classification that cannot be solved by the current recognition method?
In a lifelong learning context, new performance criteria are needed. 
For instance: where in the class space will additional labeling has the
largest beneficial training effects?

\begin{figure}[h]
\centerline{\includegraphics[width=9cm]{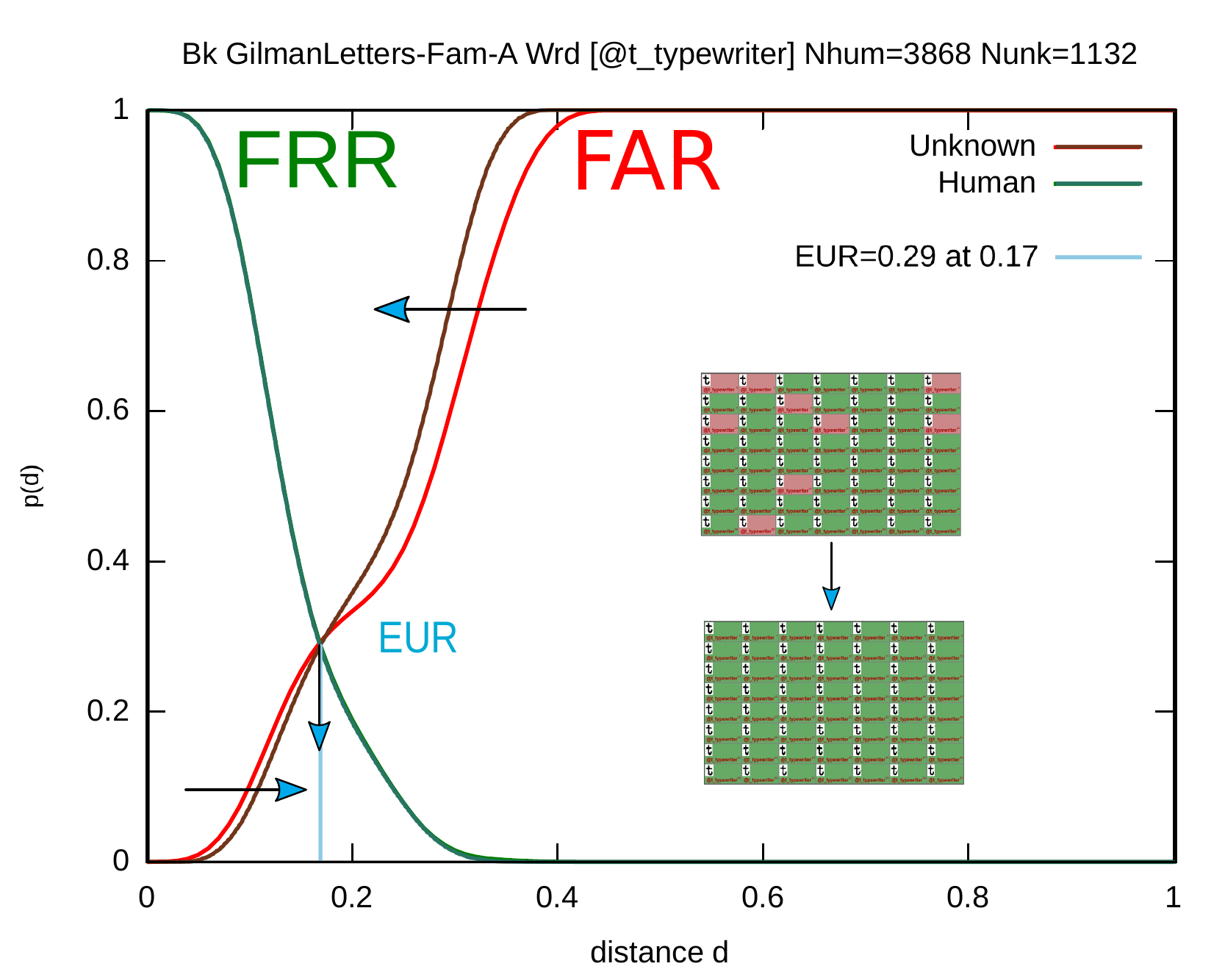}}
\caption{Schematic overview of the effects of interactive labeling on the False-Reject Rate (FRR) and
False-Accept Rate (FAR) curves for a class, in this example a typewritten letter 't'. By labeling a few
candidates (light colored) in the hitlist, they become part of the training set and have an effect on the
classification performance as indicated by the arrows. The samples and curves are an actual example
of the consequences of adding nine new instances for the letter 't'. The Monk collection does not
only contain handwriting: methods are also used for other classes of shapes. Equal Uncertainty Rate 
(EUR) is used instead of EER because the class labels of non-target patterns, i.e., the samples 
under the FAR curve, are unknown in open-set data mining.} 
\label{fig:FARcurve}
\end{figure}

An example of a new concept that was developed in this respect is the
definition of the EUR, Equal Uncertainty Rate as a replacement for the
EER, Equal Error Rate.  In case of ranking instances of a single class,
the labeled training examples are the basis for the computation of the
False Reject Rate (FRR) curve.  However, its counterpart, the False
Acceptance Rate is less clear, since it is not known whether residual
instances in a hit list are of the target class or not.  Still, the EUR
provides a good insight in the quality of instances as regards the
likelihood that they are a member of the target class or not.  By
attracting the user to those classes where FRR/FAR curves indicate a
good separating performance, human efforts are directed to where
fruitful harvest are to be expected in the mining process. 
Figure~\ref{fig:FARcurve} shows the effect of labeling on the
False-Reject Rate and False-Accept Rate curves.  The Equal Uncertainty
rate should go down.  In this real example, we see that the FRR rate
(green curve, representing the target class) remains about the same, as
well as the EUR value, but the False-Acceptance Rate curve (red,
non-target samples) is becoming steeper.  In lifelong machine learning,
it is the goal to find the classes where these improvements are largest. 
In the Monk architectures, dedicated neural networks are trained to
predict which classes give the highest improvement upon labeling.  The
users can then be attracted to these 'prospects', to confirm or
disconfirm the label values for those classes.  So basically, this
constitutes a selective-attention mechanism of the learning system,
answering the question: 'Where is the action?' in the learning process. 

%

During the writing of this chapter, a large-scale field test was performed, comparing a traditional
method and deep convolutional networks~\citep{MonkFieldTest2019}. The goal was to perform unmonitored end-to-end 
training of word, shape and character images in several hundreds of books with widely differing styles. 
Computing, using several GPUs, lasted from January 2019 to the end of April 2019.
Although deep end-to-end training was able to obtain ~95\% word accuracy in many books, the training
process failed several times for problems with more than 1000 classes in one-hot encoded classification,
even with a fairly small number of hidden units in the pre-final layer (150). Due to the fact that nearest-mean
classification using bags of visual words is not limited by the number of (lexical) classes, the average
performance over all books was higher for this traditional method (BOVW 87\% vs 83\% when failed CNN trainings were
included). As we have
shown~\citep{ShengOpenSet2018}, the problem of high numbers of classes can be solved using attribute learning, 
but this is a form of handcrafting of an output representation that would be required for different scripts and 
languages. For a system such as Monk, a pragmatic approach is chosen: For each book, the best-performing method 
can be selected. 

\section{Compositionality}

\begin{figure}[h]
\centerline{\includegraphics[width=10cm]{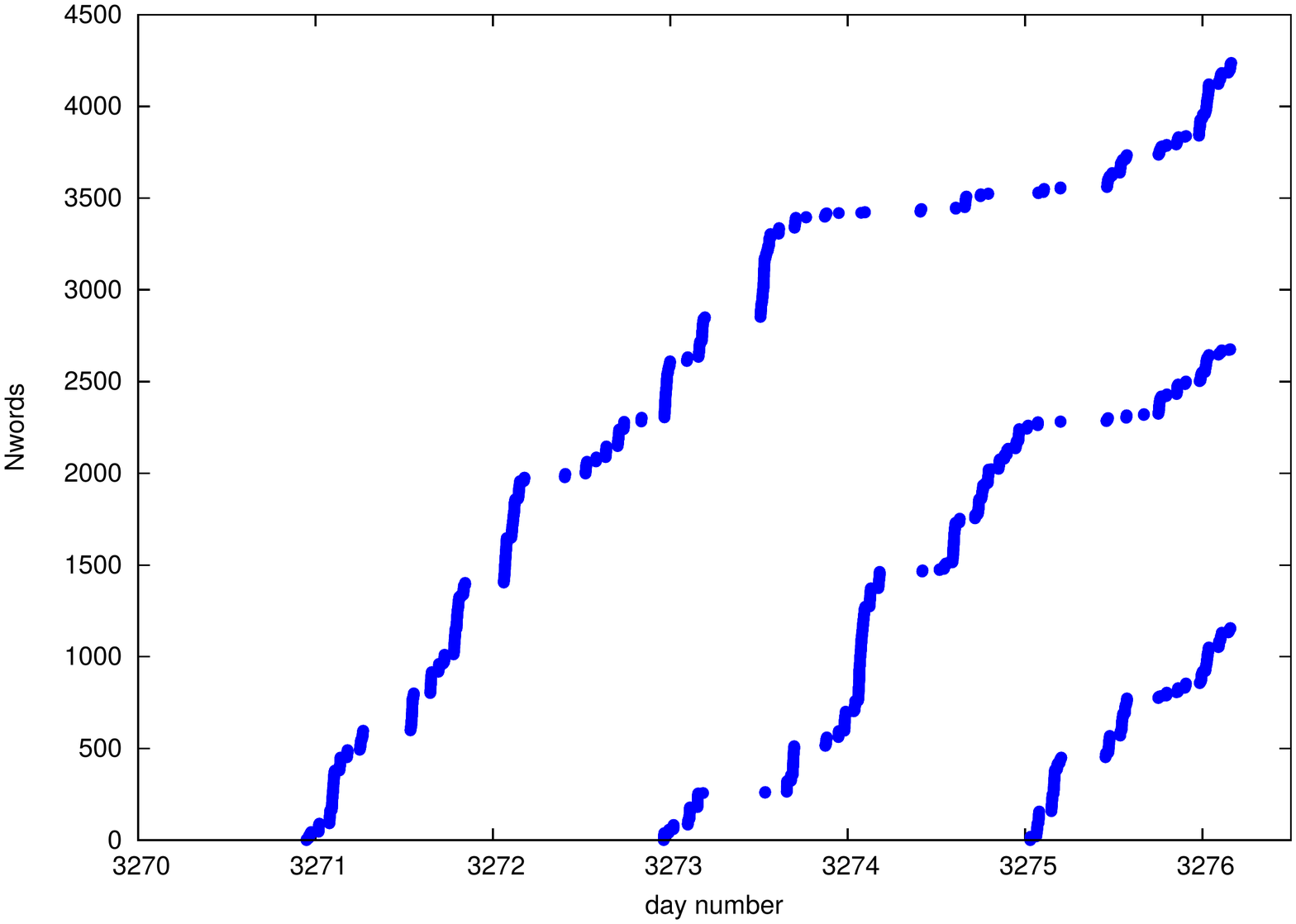}}
\medskip
\centerline{\includegraphics[width=10cm]{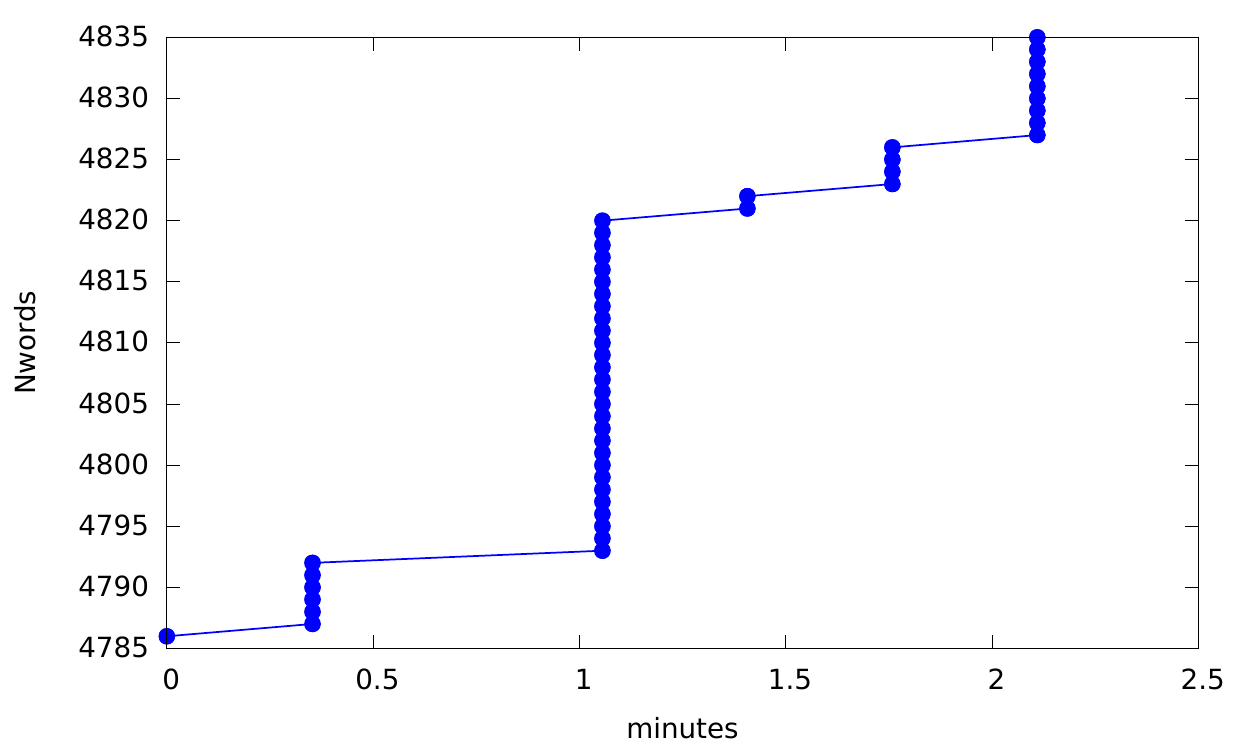}}
\caption{Examples of label harvesting over time. Top: three books of the
same collection are ingested. Note non-linear, steep speedups at some
points due to hit-list based label confirmation.
Bottom: zooming in on the growth curves show this 'snow-ball' effect
at a smaller time scale: Twenty to thirty labels are added within a minute by a single
user.}
\label{fig:Harvest}
\end{figure}

The whole-word based approach has as advantage that it
exploits the redundance of shapes and has a diminished
reliance on the well-formedness of individual character
shapes. The disadvantage is that there is a reduced
exploitation of the presence of stable pattern fractions
corresponding to letters and syllables. In the HMM-era
this was addressed by 'state sharing'. In the modern
recurrent neural networks (LSTM,BLSTM,MDLSTM), the
training process will sort this out, automatically.
However, this still requires a sufficient regularity of
character shapes to be successful, and many labeled
examples are needed, for instance 2000 lines of transcribed
text. This allows for compositionality and proper handling
of words that were never seen ('out of lexicon' recognition
rate. However, not all collections lend themselves to
the use of LSTM, because spatial invariance is not its
strongest asset and is better dealt with using CNNs.
Whole-word based approaches will be able to handle unseen
classes, if attribute schemes are used for constructing
class vector representations as targets for neural-network
output.
What is most important for a data-mining framework, is that
the harvest of labels keeps the users motivated.
Figure~\ref{fig:Harvest} shows the number of elicited labels evolving
over time. 
same collection are ingested. The growth curves show non-linear speedup at
points where labeling is facilitated due to the collaboration between man and 
machine. A detailed view on the growth curves shows the 'snow-ball' effect
at a smaller time scale: thirty labels are added within a minute by a single
user. Therefore, labeling of instances in hit lists can be 'faster than linear',
i.e., faster than begin-to-end, word-by-word transcription.

\clearpage

\section{Conclusion}

In this chapter, an overview was given of design considerations, practical
solutions and problems encountered around a large-scale multi-user e-Science
service for text recognition and retrieval in a variety of historical-document 
collections. The focus was more on text recognition than on document dating
and writer identification algorithms, which have been published elsewhere.
The major contributions of this chapter concern:

\begin{itemize}
\item A message to the machine-learning community that the current habits of
algorithm evaluation entail the risk of a narrow view on pattern diversity in
and preprocessing problems in real manuscript collections;

\medskip

\item Contrary to a separate laboratory and operational stage, continous  
improvement requires lifelong learning, in an integrated manner, also called
'persistent cognition';

\medskip

\item An active label-harvesting engine that is running in autonomous mode
needs new predictive algorithms to direct the investment of computational
resources and human labeling labour to the data portions where the most
attractive harvest prospects are to be expected. Such prospects are located
on the fringe of a training front;

\medskip

\item When critical numbers of labels are reached, the label harvesting enters
into a cascade of avalanche mode, such that ranked hit lists can be easily 
labeled, until the next performance asymptote is reached;

\medskip

\item The necessity for massive numbers of labels constitutes an intrinsic limitation
of current deep-learning methods. The ball-park principle allows for exploitation
of different types of methods which produce good results with a few or no labels,
and then progressing to more advanced methods as more labeled instances are collected
over time.
\end{itemize}

At this point, it also becomes important to add critical remarks. As a general method,
a whole-word based method is very attractive. The task is very similar to general
image retrieval, where a wide spectrum of algorithms is available to process greyscale
and color images. In Monk, a single convenient general-purpose classifier is used to
this effect. However, if individual characters actually {\em are present} and sufficiently
regular, character-based methods and recurrent networks become applicable. In the methods
described in this chapter, linguistic post processing did not play a role, because in 
most cases there is no suitable linguistic corpus. Again, however, if such a corpus
actually {\em does exist}, it would be suboptimal not to use it. Current projects
around the Monk system are directed at incorporation of linguistic and semantic resources
to improve the classification accuracy. An intrinsic problem in document retrieval concerns
the statistical distribution of terms: Many interesting target terms will be located in
the long tail of the word distribution. A low frequency of presence implies a limited 
number of training examples. Also for these cases, letter or character based methods will
be very important. An additional solution for this problem is given by the attribute-based
methods which allow for zero-shot learning. Finally, a few words need to be spent concerning
alternative projects, such as Transkribus~\citep{Transkribus2017} and READ. For historical reasons and due to 
different funding sources, the development of the Monk system took place in relative
isolation. Still, a considerable number of users from the humanities are interested in the
type of solutions provided by Monk. From their point of view, it would be very desirable if
they could benefit from the advances in each of the different approaches. As an example,
transcription data collected in Transkribus can be used to train recognizers in the Monk
system. Alternatively, text-recognition results on word images could be cross checked by 
dedicated word classification systems on manuscript-processing e-Science servers, world wide,
including those from READ, Monk and other projects. The ultimate goal would be to realize 
the type label-harvesting avalanches reported here, but over a wide range of document collections. 
In some areas, such as medieval European script styles, such a critical mass is about to be reached
in the coming few years. The costs involved in maintaining large-scale computing and storage
infrastructures will force all parties to cooperate. A good example is given by the astronomers,
who have a similarly long-term perspective as the archives and humanities scholars, but who also
have the tenacity to procure and maintain such long-term e-Science services~\citep{AstroWise2017}.

%










\bibliographystyle{ws-book-har}    
\bibliography{ws-book-Schomaker}      

\begin{thebibliography}{28}
\newcommand{\enquote}[1]{#1}
\providecommand{\natexlab}[1]{#1}
\providecommand{\url}[1]{\texttt{#1}}
\providecommand{\urlprefix}{URL }
\providecommand{\eprint}{eprint }
\expandafter\ifx\csname urlstyle\endcsname\relax
  \providecommand{\doi}[1]{doi:\discretionary{}{}{}#1}\else
  \providecommand{\doi}{doi:\discretionary{}{}{}\begingroup
  \urlstyle{rm}\Url}\fi

\bibitem[{Bluche(2016)}]{Bluche2016e}
Bluche, T. (2016). \enquote{Joint line segmentation and transcription for
  end-to-end handwritten paragraph recognition.} in \emph{29th Conference on
  Neural Information Processing Systems (NIPS)}.

\bibitem[{Bulacu \emph{et~al.}(2009)Bulacu, Brink, Zant and
  Schomaker}]{BulacuAugmentation2009}
Bulacu, M., Brink, A., Zant, T.,  and Schomaker, L. (2009).
  \enquote{Recognition of handwritten numerical fields in a large single-writer
  historical collection,} in \emph{10th International Conference on Document
  Analysis and Recognition}, pp. 808--812, \doi{10.1109/ICDAR.2009.8}.

\bibitem[{Chalmers and Edwards(2017)}]{Chalmers2017}
Chalmers, M.~K. and Edwards, P.~N. (2017). \enquote{Producing "one vast index":
  Google book search as an algorithmic system,} \emph{Big Data \& Society}
  \textbf{4}, 2, p. [unpaginated], \doi{10.1177/2053951717716950}.

\bibitem[{Chanda \emph{et~al.}(2018{\natexlab{a}})Chanda, Okafor, Hamel,
  Stutzmann and Schomaker}]{TappedFeatures2018}
Chanda, S., Okafor, E., Hamel, S., Stutzmann, D.,  and Schomaker, L.
  (2018{\natexlab{a}}). \enquote{Deep learning for classification and as
  tapped-feature generator in medieval word-image recognition,} in \emph{13th
  IAPR International Workshop on Document Analysis Systems (DAS)} (IEEE), pp.
  217--222, \doi{10.1109/DAS.2018.82}.

\bibitem[{Chanda \emph{et~al.}(2018{\natexlab{b}})Chanda, Pal, Schomaker,
  Chakraborty and Basak}]{ChandaLineSegmentation2018}
Chanda, S., Pal, U., Schomaker, L., Chakraborty, A.,  and Basak, S.
  (2018{\natexlab{b}}). \enquote{Text line segmentation and background filling
  in historical document images using grayscale information,} in
  \emph{[Submitted]} (IEEE (The Institute of Electrical and Electronics
  Engineers)), p. [unpaginated], \doi{doi}.

\bibitem[{Fischer \emph{et~al.}(2012)Fischer, Keller, Frinken and
  Bunke}]{GW2012}
Fischer, A., Keller, A., Frinken, V.,  and Bunke, H. (2012).
  \enquote{Lexicon-free handwritten word spotting using character hmms,}
  \emph{Pattern Recognition Letters} \textbf{33}, 7, pp. 934--942.

\bibitem[{Gurjar \emph{et~al.}(2018)Gurjar, Sudholt and Fink}]{Gurjar2018-LDR}
Gurjar, N., Sudholt, S.,  and Fink, G.~A. (2018). \enquote{{Learning Deep
  Representations for Word Spotting Under Weak Supervision},} in \emph{Proc.
  Int. Workshop on Document Analysis Systems} (Vienna, Austria).

\bibitem[{{He} and {Schomaker}(2018)}]{ShengOpenSet2018}
{He}, S. and {Schomaker}, L. (2018). \enquote{{Open Set Chinese Character
  Recognition using Multi-typed Attributes},} \emph{ArXiv e-prints}
  \href{http://arxiv.org/abs/1808.08993}{\UrlFont{arXiv:1808.08993 [cs.CV]}}.

\bibitem[{Hinton \emph{et~al.}(2012)Hinton, Srivastava, Krizhevsky, Sutskever
  and Salakhutdinov}]{Dropout2012}
Hinton, G.~E., Srivastava, N., Krizhevsky, A., Sutskever, I.,  and
  Salakhutdinov, R. (2012). \enquote{Improving neural networks by preventing
  co-adaptation of feature detectors,} \emph{CoRR} \textbf{abs/1207.0580},
  \href{http://arxiv.org/abs/1207.0580}{\UrlFont{arXiv:1207.0580}},
  \url{http://arxiv.org/abs/1207.0580}.

\bibitem[{Holley(2009)}]{Holley2009}
Holley, R. (2009). \enquote{How good can it get? nalysing and improving ocr
  accuracy in large scale historic newspaper digitisation programs,}
  \emph{D-Lib Magazine} \textbf{15}, 3/4, p. [unpaginated].

\bibitem[{Kahle \emph{et~al.}(2017)Kahle, Colutto, Hackl and
  M{\"{u}}hlberger}]{Transkribus2017}
Kahle, P., Colutto, S., Hackl, G.,  and M{\"{u}}hlberger, G. (2017).
  \enquote{Transkribus - {A} service platform for transcription, recognition
  and retrieval of historical documents,} in \emph{1st International Workshop
  on Open Services and Tools for Document Analysis, 14th {IAPR} International
  Conference on Document Analysis and Recognition, OST@ICDAR 2017, Kyoto,
  Japan, November 9-15, 2017}, pp. 19--24, \doi{10.1109/ICDAR.2017.307},
  \url{https://doi.org/10.1109/ICDAR.2017.307}.

\bibitem[{Klijn(2008)}]{Klijn2008}
Klijn, E. (2008). \enquote{The current state of art in newspaper digitisation.
  a market perspective.} \emph{D-Lib Magazine} \textbf{14}, 1/2, p.
  [unpaginated].

\bibitem[{Larochelle \emph{et~al.}(2008)Larochelle, Erhan and
  Bengio}]{ZeroShot2008}
Larochelle, H., Erhan, D.,  and Bengio, Y. (2008). \enquote{Zero-data learning
  of new tasks,} in \emph{Proceedings of the Twenty-Third AAAI Conference on
  Artificial Intelligence}, pp. 646--651.

\bibitem[{LeCun \emph{et~al.}(1998)LeCun, Bottou, Bengio and
  Haffner}]{MNIST1998}
LeCun, Y., Bottou, L., Bengio, Y.,  and Haffner, P. (1998).
  \enquote{Gradient-based learning applied to document recognition,}
  \emph{Proceedings of the IEEE} \textbf{86}, pp. 2278--2324,
  \url{https://en.wikipedia.org/wiki/MNIST_database}, leCun, Yann; Corinna
  Cortes; Christopher J.C. Burges. MNIST handwritten digit database.

\bibitem[{Marti and Bunke(2002)}]{IAM2002}
Marti, U. and Bunke, H. (2002). \enquote{The iam-database: An english sentence
  database for off-line handwriting recognition.} \emph{Int. Journal on
  Document Analysis and Recognition} \textbf{5}, pp. 39 -- 46,
  \url{http://www.fki.inf.unibe.ch/databases/iam-handwriting-database}.

\bibitem[{Rath and Manmatha(2007)}]{Rath2007}
Rath, T.~M. and Manmatha, R. (2007). \enquote{Word spotting for historical
  documents,} \emph{International Journal of Document Analysis and Recognition
  (IJDAR)} \textbf{9}, 2, pp. 139--152, \doi{10.1007/s10032-006-0027-8},
  \url{https://doi.org/10.1007/s10032-006-0027-8}.

\bibitem[{S\'{a}nchez \emph{et~al.}(2013)S\'{a}nchez, M\"{u}hlberger, Gatos,
  Schofield, Depuydt, Davis, Vidal and de~Does}]{Sanchez2013}
S\'{a}nchez, J.~A., M\"{u}hlberger, G., Gatos, B., Schofield, P., Depuydt, K.,
  Davis, R.~M., Vidal, E.,  and de~Does, J. (2013). \enquote{transcriptorium: A
  european project on handwritten text recognition,} in \emph{Proceedings of
  the 2013 ACM Symposium on Document Engineering}, DocEng '13 (ACM, New York,
  NY, USA), ISBN 978-1-4503-1789-4, pp. 227--228,
  \doi{10.1145/2494266.2494294},
  \url{http://doi.acm.org/10.1145/2494266.2494294}.

\bibitem[{Schomaker(2016)}]{DesignConsSchomaker2015}
Schomaker, L. (2016). \enquote{Design considerations for a large-scale
  image-based text search engine in historical manuscript collections,}
  \emph{it - Information Technology} \textbf{58}, 2, pp. 80--88,
  \url{http://www.degruyter.com/view/j/itit.2016.58.issue-2/itit-2015-0049/itit-2015-0049.xml}.

\bibitem[{Schomaker(2019)}]{MonkFieldTest2019}
Schomaker, L. (2019). \enquote{A large-scale field test on word-image
  classification in large historical document collections using a traditional
  and two deep-learning methods,} \emph{CoRR} \textbf{abs/1904.08421},
  \href{http://arxiv.org/abs/1904.08421}{\UrlFont{arXiv:1904.08421}},
  \url{http://arxiv.org/abs/1904.08421}.

\bibitem[{Sudholt and Fink(2016)}]{Sudholt2016-PAD}
Sudholt, S. and Fink, G.~A. (2016). \enquote{{PHOCNet: A Deep Convolutional
  Neural Network for Word Spotting in Handwritten Documents},} in \emph{Proc.
  Int. Conf. on Frontiers in Handwriting Recognition} (Shenzhen, China), winner
  of the IAPR Best Paper Award.

\bibitem[{Sudholt and Fink(2018)}]{Sudholt2018-ACF}
Sudholt, S. and Fink, G.~A. (2018). \enquote{{Attribute CNNs for Word Spotting
  in Handwritten Documents},} \emph{Int. Journal on Document Analysis and
  Recognition} \textbf{21}, 3, pp. 159--160.

\bibitem[{Surinta \emph{et~al.}(2014)Surinta, Karaaba, {van Oosten}, Schomaker
  and Wiering}]{SurintaLineSegAstar2017}
Surinta, O., Karaaba, M., {van Oosten}, J.-P., Schomaker, L.,  and Wiering, M.
  (2014). \enquote{A* path planning for line segmentation of handwritten
  documents,} in \emph{International Conference on Frontiers in Handwriting
  Recognition (ICFHR)} (IEEE (The Institute of Electrical and Electronics
  Engineers)), pp. 175--180, \doi{10.1109/ICFHR.2014.37}.

\bibitem[{Uchida \emph{et~al.}(2016)Uchida, Ide, Iwana and Zhu}]{Uchida2016}
Uchida, S., Ide, S., Iwana, B.~K.,  and Zhu, A. (2016). \enquote{A further step
  to perfect accuracy by training cnn with larger data,} in \emph{2016 15th
  International Conference on Frontiers in Handwriting Recognition (ICFHR)},
  pp. 405--410, \doi{10.1109/ICFHR.2016.0082}.

\bibitem[{Valentijn \emph{et~al.}(2017)Valentijn, Begeman, Belikov, Boxhoorn,
  Brinchmann, McFarland, Holties, Kuijken, {Verdoes Kleijn}, Vriend, Williams,
  Roerdink, Schomaker, Swertz, Tsyganov and {van Dijk}}]{AstroWise2017}
Valentijn, E., Begeman, K., Belikov, A., Boxhoorn, D., Brinchmann, J.,
  McFarland, J., Holties, H., Kuijken, K., {Verdoes Kleijn}, G., Vriend, W.-J.,
  Williams, O., Roerdink, J., Schomaker, L., Swertz, M., Tsyganov, A.,  and
  {van Dijk}, G. (2017). \enquote{Target and (astro-)wise technologies - data
  federations and its applications,} in \emph{Astroinformatics 2017},
  Proceedings IAU Symposium (International Astronomical Union), pp. 333--340,
  \doi{10.1017/S1743921317000254}.

\bibitem[{{van der Zant} \emph{et~al.}(2008){van der Zant}, Schomaker and
  Haak}]{Zant2008}
{van der Zant}, T., Schomaker, L.,  and Haak, K. (2008).
  \enquote{Handwritten-word spotting using biologically inspired features,}
  \emph{Ieee transactions on pattern analysis and machine intelligence}
  \textbf{30}, 11, pp. 1945--1957, \doi{10.1109/TPAMI.2008.144}.

\bibitem[{{van der Zant} \emph{et~al.}(2009){van der Zant}, Schomaker, Zinger
  and {van Schie}}]{ISR2009}
{van der Zant}, T., Schomaker, L., Zinger, S.,  and {van Schie}, H. (2009).
  \enquote{Where are the search engines for handwritten documents?}
  \emph{Interdisciplinary Science Reviews} \textbf{34}, 2-3, pp. 224--235,
  \doi{10.1179/174327909X441126}.

\bibitem[{{van Oosten} and Schomaker(2014)}]{vanOosten2014}
{van Oosten}, J.-P. and Schomaker, L. (2014). \enquote{Separability versus
  prototypicality in handwritten word-image retrieval,} \emph{Pattern
  recognition} \textbf{47}, 3, pp. 1031--1038,
  \doi{10.1016/j.patcog.2013.09.006}.

\bibitem[{Weber \emph{et~al.}(2018)Weber, Ameryan, Wolstencroft, Stork,
  Heerlien and Schomaker}]{WeberEtAl2018}
Weber, A., Ameryan, M., Wolstencroft, K., Stork, L., Heerlien, M.,  and
  Schomaker, L. (2018). \emph{Towards a Digital Infrastructure for Illustrated
  Handwritten Archives} (Springer), ISBN 978-3-319-75788-9, pp. 155--166,
  \doi{10.1007/978-3-319-75826-8_13}.

\end{thebibliography}


\printindex

\end{document}